\documentclass[10pt,twocolumn,letterpaper]{article}

\usepackage{silence}
\usepackage{wacv}
\usepackage{times}
\usepackage{epsfig}
\usepackage{graphicx}
\usepackage[font=small,labelfont=bf]{caption}
\usepackage{subcaption}
\usepackage{amsmath}
\usepackage{amssymb}
\usepackage{booktabs}

\makeatletter
\@namedef{ver@everyshi.sty}{}
\makeatother
\usepackage{tikz}

\def\wacvPaperID{1098} 

\wacvalgorithmstrack  
\wacvfinalcopy

\ifwacvfinal
\usepackage[breaklinks=true,bookmarks=false]{hyperref}
\else
\usepackage[pagebackref=true,breaklinks=true,colorlinks,bookmarks=false]{hyperref}
\fi

\usepackage{LL}

\usepackage{amsmath,amsfonts,bm}

\def\eqref#1{equation~\ref{#1}}

\def\1{\bm{1}}

\def\vmu{{\bm{\mu}}}

\def\vx{{\bm{x}}}
\def\vy{{\bm{y}}}

\DeclareMathAlphabet{\mathsfit}{\encodingdefault}{\sfdefault}{m}{sl}
\SetMathAlphabet{\mathsfit}{bold}{\encodingdefault}{\sfdefault}{bx}{n}

\usepackage{tikz}
\usepackage{pgfplots}
\usetikzlibrary{positioning, fit, shapes.geometric}
\pgfplotsset{compat=1.15}
\usepgfplotslibrary{fillbetween}

\usepackage{csvsimple}

\usepackage[group-separator={,}]{siunitx}

\newcommand\gauss[2]{1/(#2*sqrt(2*pi))*exp(-((x-#1)^2)/(2*#2^2))} 

\newcommand\gaussMixtureTwo[6]{
#5 * \gauss{#1}{#2} + #6 * \gauss{#3}{#4}
} 

\newcommand\uniformDist[2]{
1/(#2-#1)*(greater(x,#1) - greater(x,#2))
} 

\newcommand\laplaceDist[2]{
1/(2*#2)*exp(-(abs(x-#1))/(#2))
} 

\newcommand\frechet{Fr\'{e}chet}

\newcommand\MWTmath{\text{MW}_2}
\newcommand\MWT{$\text{MW}_2$}

\newcommand\resnetE{ResNet-18}
\newcommand\resnetF{ResNet-50}
\newcommand\resnext{ResNeXt-101 (32$\times$8d)}
\newcommand\inception{Inception-v3}
\newcommand{\imagenet}{ImageNet}
\newcommand{\FID}{\text{FID}}
\newcommand{\WaM}{\text{WaM}}
\newcommand{\KID}{\text{KID}}

\newcommand\colorOne{violet}
\newcommand\colorTwo{black}

\definecolor{ForestGreen}{rgb}{.133,.545,.133}
\definecolor{BurntOrange}{rgb}{.8,.333,0}

\begin{document}
\title{Evaluating generative networks using Gaussian mixtures of image features}
\author{
Lorenzo Luzi$^{1,2}$\thanks{Work done while interning at Pacific Northwest National Laboratory}, 
Carlos Ortiz Marrero$^2$, 
Nile Wynar$^2$, 
Richard G. Baraniuk$^1$, 
Michael J. Henry$^2$
\\
$^1$Rice University,
$^2$Pacific Northwest National Laboratory\thanks{Information Release Number: PNNL-SA-175469}
\\
\texttt{\{carlos.ortizmarrero,nile.wynar,michael.j.henry\}@pnnl.gov}
\\
\texttt{\{enzo,richb\}@rice.edu}
}
\maketitle
\thispagestyle{empty}
\begin{abstract}
We develop a measure for evaluating the performance of generative networks given two sets of images. A popular performance measure currently used to do this is the \frechet{} Inception Distance (FID). FID assumes that images featurized using the penultimate layer of \inception{} follow a Gaussian distribution, an assumption which cannot be violated if we wish to use FID as a metric.
However, we show that \inception{} features of the ImageNet dataset are not Gaussian; in particular, every single marginal is not Gaussian. To remedy this problem, we model the featurized images using Gaussian mixture models (GMMs) and compute the $2$-Wasserstein distance restricted to GMMs. We define a performance measure, which we call WaM, on two sets of images by using \inception{} (or another classifier) to featurize the images, estimate two GMMs, and use the restricted $2$-Wasserstein distance to compare the GMMs. We experimentally show the advantages of WaM over FID, including how FID is more sensitive than WaM to imperceptible image perturbations. By modelling the non-Gaussian features obtained from \inception{} as GMMs and using a GMM metric, we can more accurately evaluate generative network performance.
\end{abstract}
\begin{figure*}
    \captionsetup[subfigure]{justification=centering}
    \centering
    \newcommand\figWidth{0.19}
    \newcommand\xlimmax{25}
    \newcommand\xlimmaxtick{15}
    \pgfplotsset{
        histaxis/.style={
          every axis plot post/.append style={mark=none,domain=-\xlimmax:\xlimmax,samples=100},
          axis y line=none,
          axis x line*=bottom,
          xtick={-\xlimmaxtick,\xlimmaxtick},
          ymin=0.0,
          ymax=0.085,
          xmin=-\xlimmax,
          xmax=\xlimmax,
          width=3.9cm,
          xlabel={Feature value},
          }
        }
        \tikzstyle{histsLine}=[color = \colorTwo, mark = none, thick, name path = A]
        \tikzstyle{histsFill}=[color = \colorTwo, opacity=0.5]

    \begin{subfigure}[t]{\figWidth\textwidth} 
        \centering
        \begin{tikzpicture}[]
            \begin{axis}[histaxis]
                \addplot[histsLine, \colorOne] {\gauss{0}{10}};
                \addplot[draw=none,name path=B] {0};
                \addplot[histsFill, \colorOne] fill between[of=A and B,soft clip={domain=-\xlimmax:\xlimmax}];
            \end{axis}
        \end{tikzpicture}
        \caption{ \\
        \begin{tabular}{l}
        $\mu=0$
        \\
        $\sigma=10$
        \end{tabular}
        }
        \label{subfig:gaussian}
    \end{subfigure}%
    \begin{subfigure}[t]{\figWidth\textwidth} 
        \centering       \begin{tikzpicture}[]
            \begin{axis}[histaxis]
                \addplot[histsLine] {\gaussMixtureTwo
                {-4*sqrt{20}}{sqrt{40}}{sqrt{20}}{sqrt{15}}{0.2}{0.8}};
                
                \addplot[draw=none,name path=B] {0};
                \addplot[histsFill] fill between[of=A and B,soft clip={domain=-\xlimmax:\xlimmax}];
            \end{axis}
        \end{tikzpicture}
        \caption{
        \begin{tabular}{l}
        $\mu_1, \mu_2 = -8\sqrt{5}, 2\sqrt{5}$
        \\
        $\sigma_1, \sigma_2 = 2\sqrt{10}, \sqrt{15}$
        \\
        $\pi_1, \pi_2 = 0.2, 0.8$
        \end{tabular}}
        \label{subfig:GMMskew}
    \end{subfigure}
    \begin{subfigure}[t]{\figWidth\textwidth} 
        \centering       \begin{tikzpicture}[]
            \begin{axis}[histaxis]
                \addplot[histsLine] {
                \uniformDist{-sqrt(300)}{sqrt(300)}
                };
                
                \addplot[draw=none,name path=B] {0};
                \addplot[histsFill] fill between[of=A and B,soft clip={domain=-\xlimmax:\xlimmax}];
            \end{axis}
        \end{tikzpicture}
        \caption{ \\
        \begin{tabular}{l}
        $a = -10\sqrt{3}$
        \\
        $b = 10\sqrt{3}$
        \end{tabular}}
        \label{subfig:uniform}
    \end{subfigure}
    \begin{subfigure}[t]{\figWidth\textwidth} 
        \centering       \begin{tikzpicture}[]
            \begin{axis}[histaxis]
                \addplot[histsLine] {\gaussMixtureTwo
                {-1*sqrt{80}}{sqrt{20}}{sqrt{80}}{sqrt{20}}{0.5}{0.5}};
                
                \addplot[draw=none,name path=B] {0};
                \addplot[histsFill] fill between[of=A and B,soft clip={domain=-\xlimmax:\xlimmax}];
            \end{axis}
        \end{tikzpicture}
        \caption{ 
        \begin{tabular}{l}
        $\mu_1, \mu_2 = -4\sqrt{5}, 4\sqrt{5}$
        \\
        $\sigma_1, \sigma_2 = 2\sqrt{5}, 2\sqrt{5}$
        \\
        $\pi_1, \pi_2 = 0.5, 0.5$
        \end{tabular}}
        \label{subfig:GMMsymmetric}
    \end{subfigure}
    \begin{subfigure}[t]{\figWidth\textwidth} 
        \centering       \begin{tikzpicture}[]
            \begin{axis}[histaxis]
                \addplot[histsLine] {
                \laplaceDist{0}{sqrt(50)}
                };
                
                \addplot[draw=none,name path=B] {0};
                \addplot[histsFill] fill between[of=A and B,soft clip={domain=-\xlimmax:\xlimmax}];
            \end{axis}
        \end{tikzpicture}
        \caption{ \\
        \begin{tabular}{l}
        $\mu = 0$
        \\
        $b = 5\sqrt{2}$
        \end{tabular}}
        \label{subfig:laplace}
    \end{subfigure}
    \caption{The FID score between each pair of the distributions shown above is zero  although they are clearly different distributions. This is because \cref{eq:wasserstein between gaussians} is only defined for Gaussians, and FID treats any input distribution as Gaussian, even if it is not. We plot one dimensional distributions here for visualization purposes, but the FID score will remain zero even if we extend these distributions to their high dimensional isotrophic counterparts. All that is required for the FID score between two distributions to be zero is that their first two moments match.
    \cref{subfig:gaussian} is the only Gaussian distribution. \cref{subfig:GMMskew,subfig:GMMsymmetric} are Gaussian mixtures with two components, \cref{subfig:uniform} is a uniform distribution, and \ref{subfig:laplace} is a Laplace distribution.}
    \label{fig:FIDisZero}
\end{figure*}
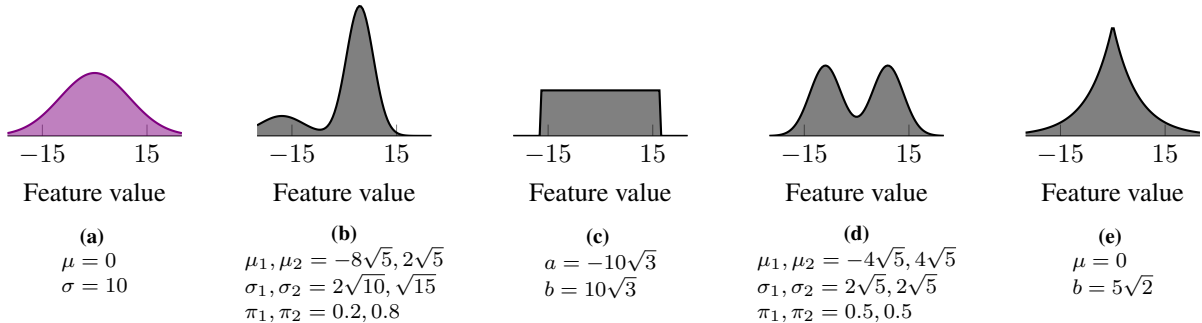
\section{Introduction }
Generative networks, such as generative adversarial networks (GANs)~\cite{goodfellow2014generative} and variational autoencoders~\cite{kingma2013auto}, model distributions implicitly by trying to learn a map from a simple distribution, such as a Gaussian, to the desired target distribution. 
Using generative networks, one can 
generate new images~\cite{biggan,styleGAN,styleGAN2,karras2017progressive,kingma2013auto}, 
superresolve images~\cite{ledig2017photo,wang2018esrgan}, 
solve inverse problems~\cite{bora2017compressed}, and perform a host of image-to-image translation tasks~\cite{isola2017image,zhu2017unpaired,zhu2016generative}. 
However, the high dimensionality of an image distribution makes it difficult to model explicitly, that is, to estimate the moments of the distribution via some parameterization.
Just estimating the covariance of a distribution requires $\frac{p(p+1)}{2}$ parameters, where $p$ is the feature dimension. 
For this reason, modelling distributions implicitly, using transformations of simple distributions, can be useful for high dimensional data. Since the generator network is typically nonlinear, the explicit form of the generated distribution is unknown. Nonetheless, these generative models allow one to sample from the learned distribution.

Because we only have access to samples from these generative networks, instead of explicit probability density functions, evaluating their performance can be difficult.
Several ways of evaluating the quality of the samples drawn from generative networks~\cite{borji2019pros} have been proposed, the most popular of which is the \frechet{} Inception distance (FID)~\cite{FID}. 
FID fits Gaussian distributions to features extracted from a set of a real images and a set of GAN-generated images.
The features are typically extracted using the \inception{} classifier~\cite{inceptionv3}.
These two distributions are then compared using the $2$-Wasserstein~\cite{villani2009optimal,villani2003topics} metric. 
While FID has demonstrated its utility in providing a computationally efficient metric for assessing the quality of GAN-generated images, our examination reveals that the fundamental assumption of FID---namely, that the underlying feature distributions are Gaussian---is invalid.
A more accurate model of the underlying features will capture a more comprehensive and informative assessment of GAN quality.

In this paper, we first show that the features used to calculate FID are not Gaussian, violating the main assumption in FID (\cref{sec:not Gaussian}). As we depict in \cref{fig:FIDisZero}, this can result in an FID value of $0$ even when the data distributions are completely different.
This happens because FID only captures the first two moments of the feature distribution and completely ignores all information present in the higher order moments. 
Thus, FID is biased toward artificially low values and invariant to information present in the higher order moments of the featurized real and generated data.

Thus, we propose a Gaussian mixture model (GMM)~\cite{finiteMixtureModels} for the features instead for several reasons.
First, GMMs can model complex distributions and capture higher order moments. In fact, any distribution can be approximated arbitrarily well with a GMM~\cite{delon2020wasserstein}.
Second, GMMs are estimated efficiently on both CPU and GPU.
Third, there exists a Wasserstein-type metric for GMMs~\cite{delon2020wasserstein} (\cref{sec:WaM}) which allows us to generalize FID. We use this newly developed metric from optimal transport to construct our generative model evaluation metric, WaM.

We show that WaM is not as sensitive to visually imperceptible noise as FID (\cref{sec:Experiments}). This is important because we do not want our evaluations metrics to vary widely between different generated datasets if we cannot visually see any difference between them. 
Since GMMs can capture more information than Gaussians, WaM more accurately identifies differences between sets of images and avoids the low score bias of FID.
We therefore reduce the issue of FID being overly sensitive to various noise perturbations~\cite{borji2019pros} by modelling more information in the feature distributions.
We test perturbation sensitivity using additive isotropic Gaussian noise and perturbed images which specifically attempt to increase the feature means using backpropagation~\cite{mathiasen2020fast}. 
The ability of WaM to model more information in the feature distribution makes it a better evaluation metric than FID for generative networks.

\section{Related work}

\label{sec:related work}

\subsection{Wasserstein distance}
A popular metric from optimal transport~\cite{villani2003topics,villani2009optimal} is the $p$-Wasserstein metric. 
Let $X$ be a Polish metric space with a metric $d$.
Given $p\in (0,\infty)$ and two distributions $P$ and $Q$ on $X$ with finite moments of order $p$, the $p$-Wasserstein metric is given by
\[
    \mcW_p(P,Q) =
    \left( \inf_{\gamma} \int_{X\times X} d(x,y)^p d\gamma(x,y) \right)^\frac{1}{p}
\]
where the infimum is taken over all joint distributions $\gamma$ of $P$ and $Q$. Different values of $p$ yield different metric properties; in image processing, the $1$-Wasserstein metric on discrete spaces is often used and called the earth mover distance~\cite{emd}. The $2$-Wasserstein metric~\cite{dowson1982frechet,olkin1982distance} is often used when comparing Gaussians since there exists a closed form solution. The formula
\begin{align}
    \label{eq:wasserstein between gaussians}
    &\mcW_2^2(\mcN(\vmu_1, \mSig_1), \mcN(\vmu_2, \mSig_2))
    =
    \| \vmu_1 - \vmu_2\|_2^2
    \\ \nonumber 
    &\hphantom{---} +
    \trace(\mSig_1) + \trace(\mSig_2) - 2 \trace\Big(  \big(\mSig_1^\frac{1}{2} \mSig_2 \mSig_1^\frac{1}{2}\big)^\frac{1}{2} \Big)
\end{align}
is used to calculate the \frechet{} Inception distance.

\subsection{FID and variants}

The \frechet{} Inception distance (FID)~\cite{FID} is a performance measure typically used to evaluate generative networks. In order to compare two sets of images, $X_1$ and $X_2$, they are featurized using the penultimate layer of the \inception{} network to get sets of features $F_1$ and $F_2$. For \imagenet{} data, this reduces the dimension of the data from $3\times 224\times 224 = $ \num{150528} to \num{2048}. These features are assumed to be Gaussian, allowing \cref{eq:wasserstein between gaussians} to be used to obtain a distance between them.

There are several ways that FID has been improved. One work has shown that FID is biased~\cite{chong2020effectively}, especially when it is computed using a small number of samples. 
They show that FID is unbiased asymptotically and show how to estimate the asymptotic value of FID to obtain an unbiased estimate.
Others have used a network different from \inception{} to evaluate data that is not from \imagenet{}; for example, a LeNet-like~\cite{lecun1989backpropagation} feature extractor can be used for MNIST.
In this work we focus on several different \imagenet{} feature extractors because of their widespread use.
Modelling \imagenet{} features has been improved due to a conditional version of FID~\cite{soloveitchik2021conditional} which extends FID to conditional distributions, and a class-aware \frechet{} distance~\cite{liu2018improved} which models the classes with GMMs. 
In this work, we do not consider conditional versions of FID, but our work can be extended to fit such a formulation in a straightforward manner. Moreover, we use GMMs over the feature space rather than one component per class as is done in the class-aware \frechet{} distance.

The kernel Inception distance~\cite{binkowski2018demystifying} is calculated by mapping the image features to a reproducing kernel Hilbert space and then using an unbiased estimate of maximum mean discrepancy to calculate a distance between sets of images. 
We compare to KID in \cref{app:KID}.

Another related metric is called WInD~\cite{WInD}. WInD uses a combination of the $1$-Wasserstein metric on discrete spaces with the $2$-Wasserstein metric on $\bbR^p$. For this reason, it is not a $p$-Wasserstein metric in $\bbR^p$ or between GMMs. For example, if $P$ and $Q$ are a mixture of Dirac delta functions, then the WInD distance between them becomes the $1$-Wasserstein distance. However, if $P$ and $Q$ are Gaussian, then the WInD distance between them becomes the $2$-Wasserstein distance.
Moreover, if $P$ and $Q$ are arbitrary GMMs, the relationship between WInD and the $p$-Wasserstein metrics is not clear.
This means that WInD can alternate between the $1$-Wasserstein and $2$-Wasserstein distance depending on the input distributions.
In this paper, we focus on using a metric which closely follows the $2$-Wasserstein distance as is currently done with FID.

\subsection{ 2-Wasserstein metric on GMMs: \texorpdfstring{\MWT{}}{MW2} }\label{sec:MW2}

A closed form solution for the $2$-Wasserstein distance between GMMs is not known. This is because the joint distribution between two GMMs is not necessarily a GMM. However, if we restrict ourselves to the relaxed problem of only considering joint distributions over GMMs, then the resulting $2$-Wasserstein distance of this new space is known.
The restricted space of GMMs is quite large, since GMMs can approximate any distribution to arbitrary precision given enough mixture components.
So given two GMMs, $P$ and $Q$, we can calculate
\[
    \MWTmath^2(P,Q) =
    \inf_{\gamma} \int_{X\times X} d(x,y)^2 d\gamma(x,y) 
\]
where the infimum is over all joint distributions $\gamma$ which are also GMMs. 
Constraining the class of joint distributions is a relaxation that has been done before~\cite{bion2019wasserstein} due to the difficulty of considering arbitrary joint distributions.
This metric, \MWT{}, appears in a few different sources in the literature~\cite{chen2016distance,chen2018optimal,chen2019aggregated} and has been studied theoretically~\cite{delon2020wasserstein}; recently, implementations of this quantity have emerged.\footnote{\url{https://github.com/judelo/gmmot}}

The practical formulation of this problem is done as follows.
Let $P = \sum_{i=1}^{K_0} \pi_i \nu_i$ and $Q = \sum_{j=1}^{K_1} \alpha_j \mu_j$ be two GMMs with  Gaussians $\nu_i, \mu_j$ for $i \in \{1,\dots, K_0\}, j \in \{1, \dots, K_1\}$. Then, we have that
\begin{equation}
    \MWTmath^2(P,Q)
    =
    \min_\gamma \sum_{ij} \gamma_{ij} \mcW_2^2(\nu_i, \mu_j)\label{MW2}
    \end{equation}
where $\gamma$ is taken to be the joint distribution over the two categorical distributions $\begin{bmatrix} \pi_1 & \hdots & \pi_{K_0}\end{bmatrix}$ and $\begin{bmatrix} \alpha_1 & \hdots & \alpha_{K_1}\end{bmatrix}$; hence, $\gamma$ in this case is actually a matrix. Thus, \MWT{} can be implemented as a discrete optimal transport plan and efficient software exists to compute this~\cite{flamary2021pot}.

\MWT{} is a great candidate for modelling the distance between GMMs for several reasons; most importantly, it is an actual distance metric. Since we are restricting the joint distribution to be a GMM, we see that \MWT{} must be greater than or equal to the $2$-Wasserstein distance between two GMMs. 
Moreover, \MWT{} clearly approximates the $2$-Wasserstein metric; there are bounds showing how close \MWT{} is to $\mcW_2$~\cite{delon2020wasserstein}. It is also computationally efficient to compute because it can be formulated as a discrete optimal transport problem, making it practical. The strong theoretical properties and computational efficiency of \MWT{} make it a prime candidate to calculate the distance between GMMs.

\section{\inception{} has Non-Gaussian features on ImageNet }
\label{sec:not Gaussian}

\subsection{Non-Gaussian features can differ and have zero FID}

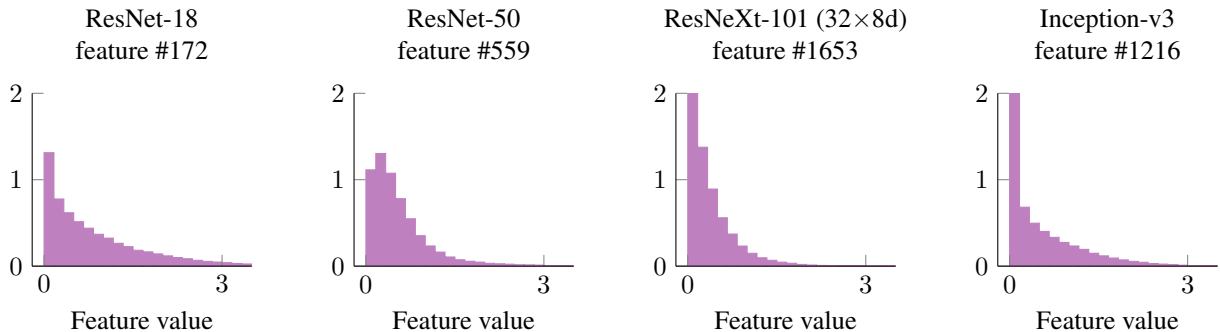
\begin{figure*}[t]
    \centering
    \newcommand\figWidth{0.24}
    \newcommand\xlimupper{3.5}
    \newcommand\xlimlower{-0.2}
    \newcommand\xlimtickupper{3}
    \newcommand\xlimticklower{0}
    \pgfplotsset{
        histaxis/.style={
          xtick={\xlimticklower,\xlimtickupper},
          ymax=2.0,
          ytick={0,1,2},
          xmin=\xlimlower,
          xmax=\xlimupper,
          width=4.5cm,
          ymin=0, 
          axis x line*=bottom,
          axis y line*=left,
          area style,
          xlabel={Feature value}
          }
        }
        \tikzstyle{histsFill}=[
                    color=\colorOne,
                    opacity=0.5, 
                    mark=no, 
                    ybar interval,
                    fill]

    \begin{subfigure}[t]{\figWidth\textwidth} 
        \centering       
        \begin{tikzpicture}[trim axis left, trim axis right]
            \begin{axis}[histaxis,
                         title={
                             \begin{tabular}{c}
                             \resnetE{}
                             \\
                             feature \#172
                             \end{tabular}
                         }]
            \addplot+[ histsFill] 
                table [col sep=comma] {csv/hist_resnet18_marginal.csv};
            \end{axis}
        \end{tikzpicture}
    \end{subfigure}
    \begin{subfigure}[t]{\figWidth\textwidth} 
        \centering       
        \begin{tikzpicture}[trim axis left, trim axis right]
            \begin{axis}[histaxis,
                         title={
                             \begin{tabular}{c}
                             \resnetF{}
                             \\
                             feature \#559
                             \end{tabular}
                         }]
            \addplot+[histsFill] 
                table [col sep=comma] {csv/hist_resnet50_marginal.csv};
            \end{axis}
        \end{tikzpicture}
    \end{subfigure}
    \begin{subfigure}[t]{\figWidth\textwidth} 
        \centering       
        \begin{tikzpicture}[trim axis left, trim axis right]
            \begin{axis}[histaxis,
                         title={
                             \begin{tabular}{c}
                             \resnext{}
                             \\
                             feature \#1653
                             \end{tabular}
                         }]
            \addplot+[histsFill] 
                table [col sep=comma] {csv/hist_resnext101_32x8d_marginal.csv};
            \end{axis}
        \end{tikzpicture}
    \end{subfigure}
    \begin{subfigure}[t]{\figWidth\textwidth} 
        \centering
        \begin{tikzpicture}[trim axis left, trim axis right]
            \begin{axis}[histaxis,
                         title={
                             \begin{tabular}{c}
                             \inception{}
                             \\
                             feature \#1216
                             \end{tabular}
                         }]
            \addplot+[histsFill] 
                table [col sep=comma] {csv/hist_inception_v3_marginal.csv};
            \end{axis}
        \end{tikzpicture}
    \end{subfigure}

    \caption{Histograms showing non-Gaussianity of randomly chosen features from the \imagenet{} validation dataset featurized by \resnetE{}, \resnetF{}, \resnext{}, and \inception{}.
    They are non-negative because these features are passed through a ReLU layer and then average pooled; for this reason, we have a spike around $0$. These histograms are empirical distributions. 
    }
    \label{fig:feature marginals}
\end{figure*}

The calculation of FID assumes that features from the penultimate layer of \inception{}~\cite{inceptionv3} are Gaussian. This layer average pools the outputs of several convolutional layers which are rectified via the ReLU activation. 
Though an argument can be made for why the preactivations of the convolutional layers are Gaussian (using the central limit theorem), the rectified and averaged outputs are not. In fact, they are likely to be averages of rectified Gaussians~\cite{beauchamp2018numerical}. Although these features are high dimensional and cannot be visualized, we plot the histograms of a randomly selected feature extracted with \inception{}, \resnetE{}, \resnetF{}, and \resnext{} in \cref{fig:feature marginals}. We construct these histograms using the \num{50000} images in the \imagenet{} validation dataset. We see that none these randomly selected features appear Gaussian.

If the Gaussian assumption of FID is false, one can achieve low FID values while having drastically different distributions, as shown on \cref{fig:FIDisZero}. This is true in part because FID only considers the first two moments of the distributions being compared; differences in skew and higher order moments are not taken into account in the FID calculation. This can cause FID to be extremely low when the distributions being compared are quite different.

\subsection{\imagenet{} features are not Gaussian}

Testing if \inception{} features are Gaussian is not trivial because they are 2048-dimensional. Even if each marginal distribution appears Gaussian, we cannot be sure that the joint distribution is Gaussian. However, if the marginals are not Gaussian, this implies that original distribution is not Gaussian. Therefore, we conducted a series of Kolmogorov–Smirnov hypothesis tests \cite{kstest}, a statistical nonparametric goodness-of-fit test that verifies whether an underlying probability distribution, in our case the marginals, differs from a hypothesized distribution, a Gaussian distribution.

We calculated features from the entire \imagenet{} validation dataset using \resnetE{}, \resnetF{}, \resnext{}, and  \inception{}. For each set of features, we then tested each marginal using the Kolmogorov–Smirnov tests with the hypothesis that the features come from a normal distribution. Using a $p$-value of $0.01$, the test found that $100\%$ of the marginals fail to pass the hypothesis. This confirms, with high certainty, that neither the marginals nor the whole feature distribution is Gaussian.

Since the features of \inception{} are not Gaussian, we have a few options. The first option is to use features before the average pooling layer and ReLU operation because these features may actually be Gaussian. 
However, these features are extremely high dimensional ($64 \times 2048=$ \num{131072}) and thus very hard to estimate accurately.
Alternatively, we can remove the ReLU operation, but this would distort the features by removing the nonlinearity that is so critical to deep networks.
Another option we have is to use a different network for feature extraction; however, most networks which perform very well on \imagenet{} have high dimensionality convolutional features followed by ReLU and average pooling, e.g., \resnetE{}, \resnetF{}, and \resnext{}.
Moreover, trying to obtain Gaussian features is not a general solution because even if the training data has Gaussian features, new data may not.
Therefore, we decided to model these non-Gaussian features using Gaussian mixture models which can capture information past the first two moments of a distribution.

\section{WaM --- Model details}
\label{sec:WaM}

\subsection{A Gaussian mixture model can learn more complex distributions}

In this work we use the Gaussian mixture model (GMM) to model non-Gaussian features. GMMs are a generalization of Gaussian distributions (i.e., when the number of components equal \num{1}) and hence we can generalize FID using the formulas discussed in Section \ref{sec:MW2}. Moreover, any distribution can be approximated to arbitrary precision using a GMM~\cite{delon2020wasserstein}.
Estimation of GMM parameters are also computationally efficient and have been studied thoroughly~\cite{bishop2006pattern,finiteMixtureModels}.
Most importantly, we can calculate the distance between GMMs using \eqref{MW2}.

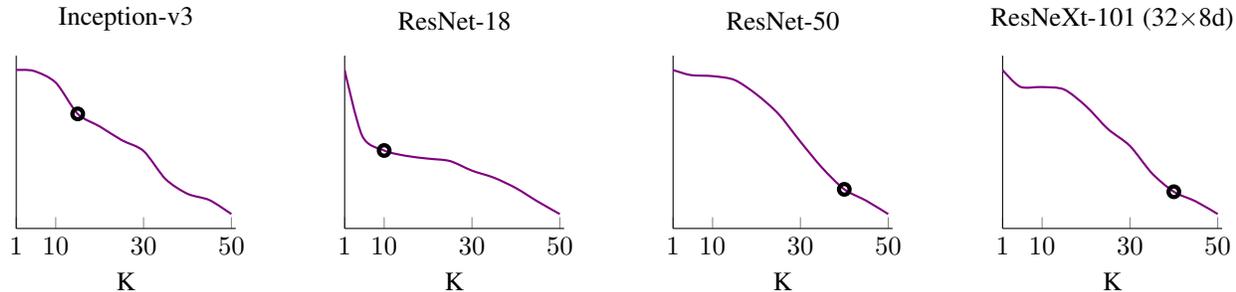
\begin{figure*}
    \centering
    \newcommand\figWidth{0.245}
    \newcommand\xlimupper{51}
    \newcommand\xlimlower{1}
    \pgfplotsset{
        axisStyleLine/.style={
          xtick={1,10,30,50},
          xmin=\xlimlower,
          xmax=\xlimupper,
          ymajorticks=false,
          width=4.5cm,
          axis x line*=bottom,
          axis y line*=left,
          xlabel={K},
          }
        }
        \tikzstyle{plotStyleLine}=[
                    color=\colorOne, 
                    mark={},
                    smooth, 
                    thick]
        \tikzstyle{plotStylePoint}=[
                    color=\colorTwo, 
                    only marks,
                    mark=o,
                    ultra thick,
                    mark size=2pt
                    ]
    \begin{subfigure}[t]{\figWidth\textwidth} 
        \centering
        \begin{tikzpicture}
            \begin{axis}[axisStyleLine, title={\inception{}}]
            \addplot+[plotStyleLine] 
                table [col sep=comma] {csv/k_inception_v3.csv};
            
            \addplot[plotStylePoint] coordinates {(15,423167287.7144545)};
            \end{axis}
        \end{tikzpicture}
    \end{subfigure}
    \begin{subfigure}[t]{\figWidth\textwidth} 
        \centering       
        \begin{tikzpicture}
            \begin{axis}[axisStyleLine, title={\resnetE{}}]
            \addplot+[ plotStyleLine] 
                table [col sep=comma] {csv/k_resnet18.csv};
                
            \addplot[plotStylePoint] coordinates {(10,91484099.8113528)};
            \end{axis}
            
        \end{tikzpicture}
    \end{subfigure}
    \begin{subfigure}[t]{\figWidth\textwidth} 
        \centering       
        \begin{tikzpicture}
            \begin{axis}[axisStyleLine, title={\resnetF{}}]
            \addplot+[plotStyleLine] 
                table [col sep=comma] {csv/k_resnet50.csv};
                
            \addplot[plotStylePoint] coordinates {(40,-105748818.10455507)};
            \end{axis}

        \end{tikzpicture}
    \end{subfigure}
    \begin{subfigure}[t]{\figWidth\textwidth} 
        \centering       
        \begin{tikzpicture}
            \begin{axis}[axisStyleLine, title={\resnext{}}]
            \addplot+[plotStyleLine] 
                table [col sep=comma] {csv/k_resnext101_32x8d.csv};
                
            \addplot[plotStylePoint] coordinates {(40,45519823.75720617)};
            \end{axis}

        \end{tikzpicture}
    \end{subfigure}

    \caption{AIC curves for features used for picking the number of mixture components $K$. We choose $K=10$ for \resnetE{}, $K=40$ for both \resnetF{} and \resnext{}, and $K=15$ and \inception{}.  }
    \label{fig:choosing k}
\end{figure*}

Before modelling the image features with GMMs, we transform them using a simple element-wise natural logarithm transformation; i.e., $\vx' = \text{ln}(\vx)$ for features $\vx$. This squashes the peak and make the data easier to model~\cite{finiteMixtureModels} although it is still not easily modeled by just one Gaussian distribution.

We calculate our performance metric for generative models by using the \MWT{}~\cite{delon2020wasserstein} metric for GMMs on GMMs estimated from extracted features of images.
The procedure is summarized as follows. We first pick a network, such as \inception{}, to calculate the features. These features are then used to estimate a GMM with $K$ components. 
We do this for real data and for generated data.
We then calculate the FID of each combination of components, one from the real data GMM and one from the generated data GMM. 
Then, we solve a discrete optimal transport problem using the $2$-Wasserstein distance squared as the ground distances to obtain WaM.
We use $n=$ \num{50000} samples because this was shown to be an approximately unbiased~\cite{chong2020effectively} estimate of FID.
We call our metric \textbf{WaM} since it is a \textbf{Wa}sserstein-type metric on G\textbf{M}Ms of image features.

We fit the GMM to the data using the expectation maximization algorithm implemented in scikit-learn~\cite{sklearn} and pycave\footnote{\url{https://github.com/borchero/pycave}}. We model the features with full covariance matrices so that we are truly generalizing FID. 
One can fit diagonal or spherical covariance matrices if speed is required, but this will yield simpler GMMs.
We considered several GPU implementations of GMM fitting instead of the scikit-learn CPU implementation. However, the sequential nature of the expectation maximization algorithm caused the run times to be similar for GPU and CPU algorithms.

\subsection{Using different networks}

In addition to using \inception{} for feature extraction, we also use \resnetE{}, \resnetF{}, and \resnext{} trained on~\imagenet{}.
For each network, we use the penultimate layer for feature extraction, as was done originally for \inception{}.
We use \resnetE{} because its features are only \num{512}-dimensional and hence can be calculated faster than \inception{}. \resnetF{} performs better than \resnetE{} and so we included it in some of our experiments. Finally, \resnext{} achieves the highest accuracy in the \imagenet{} classification task of all the pretrained classifiers on Pytorch~\cite{pytorch}.

\subsection{Picking \texorpdfstring{$K$}{K} and fitting the GMM}

When modelling features, we must pick the number of components we choose to have in our GMM. If we pick $K=1$ (and use \inception{} as our feature extractor), then we just calculate FID. The more components we pick, the better our fit will be. However, if we pick $K$ to be too large, such as $K\geq N$, then we may overfit in the sense that we can have each component centered around single data points. This is clearly not desirable, so we fit all of our GMMs with a maximum of $K=50$ components.

We use the Akaike information criterion (AIC) to choose $K$ since likelihood criteria are well suited for density estimation~\cite{finiteMixtureModels} as compared with cross validation for clustering.
However, calculating AIC for multiple components will take significant computation time and power if done every time one wants to calculate WaM. For this reason, we pick a specific $K$ based on the \imagenet{} validation set. A value for $K$ which models the \imagenet{} validation dataset well should be a good $K$ for modelling similar image datasets. As shown in \cref{fig:choosing k}, the AIC curves have varying shapes.
We use the kneed method~\cite{satopaa2011finding} for our choice of $K$ (using $S=0.5$ in the official implementation~\footnote{\url{https://github.com/arvkevi/kneed}}) for the \resnetE{}, \resnetF{}, \resnext{}, and \inception{} features. In the calculation of the knee, we ignore the first few points of the plots because desirable knees lie in the convex part of the plot, not the concave part.

Since GMMs have more parameters, they are computationally more expensive to train than simply modelling the data as a Gaussian. However, we use GMM training procedures that  take advantage of GPU parallelization\footnote{\url{https://github.com/borchero/pycave}}. As shown on \cref{tab:times}, fitting a $20$ component GMM only takes approximately 100 seconds and calculating WaM takes an additional  60 seconds. In these calculations, we compare to a fixed reference dataset with precalculated parameters as is typically done. From empirical observations, FID takes about $20$ seconds to compute, making WaM only $140$ seconds, or about $2$ minutes, slower.

\begin{table}[!h]
    \centering
    \bgroup
    \setlength\tabcolsep{4pt}
    \begin{tabular}{lcccccc}
        \toprule
        \textbf{k}
        &
        5
        &
        10
        &
        15
        &
        20
        &
        25
        &
        30
        \\ \midrule
        GMM Fitting
        & 
        $51.1$
        &
        $83.3 $
        &
        $78.3 $
        &
        $99.8 $
        &
        $143.2 $
        &
        $139.2 $
        \\ 
        WaM Calc & 
        $17.4$
        &
        $32.2$
        &
        $47.1$
        &
        $60.3$
        &
        $74.6$
        &
        $86.9$
        \\ \bottomrule
    \end{tabular}
    \egroup
    \caption{ Average number of seconds it takes to fit a GMM and calculate WaM on one GPU. This makes WaM approximately 2 minutes slower than FID.}
    \label{tab:times}
\end{table}

\section{Experiments }
\label{sec:Experiments}
In these experiments we find that WaM performs  better than both FID and KID. The KID experiments are in \cref{app:KID}.
\begin{figure*}
    \centering
    \def\figWidth{0.24}
    \bgroup
    \setlength\tabcolsep{2pt}
    \begin{tabular}{cccc}
        \toprule
        \textbf{Original (BigGAN)}
        &
        \textbf{Perturbed (BigGAN)}
        &
        \textbf{Original (ImageNet)}
        &
        \textbf{Perturbed (ImageNet)}
        \\
        \midrule
        $\!\begin{aligned}
        \FID &= 55.71 \\
        \WaM^2 &= 378.37 \end{aligned}$
        &
        $\!\begin{aligned}
        \FID &= 154.19 \\
        \WaM^2 &= 424.29 \end{aligned}$
        &
        $\!\begin{aligned}
        \FID &= 3.66 \\
        \WaM^2 &= 237.05 \end{aligned}$
        &
        $\!\begin{aligned}
        \FID &= 46.63 \\
        \WaM^2 &= 280.02 \end{aligned}$
        \\
        \includegraphics[width=\figWidth\textwidth]{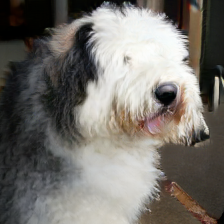}
        &
        \includegraphics[width=\figWidth\textwidth]{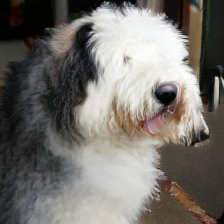}
        &
        \includegraphics[width=\figWidth\textwidth]{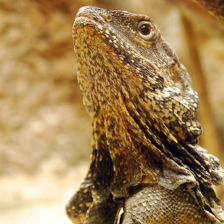}
        &
        \includegraphics[width=\figWidth\textwidth]{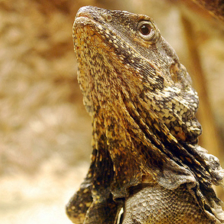}
        \\
        \multicolumn{2}{c}
        {
        $\!\begin{aligned}
        R_\FID &= 2.77 \\
        R_\WaM &= 1.12 \\
        R &= 2.47 \end{aligned}$}
        &
        \multicolumn{2}{c}
        {$\!\begin{aligned}
        R_\FID &= 12.74 \\
        R_\WaM &= 1.18 \\
        R &= 10.78 \end{aligned}$}
        \\
        \bottomrule
    \end{tabular}
    \egroup
    \caption{ Samples of images showing targeted perturbations which target the feature means, as defined on \cref{eq:meanPert}. 
    The two original images above are randomly selected from a set of \num{50000} images generated by BigGAN and a set of \num{50000} images of the \imagenet{} validation dataset. 
    We cannot visually perceive the difference between the original and perturbed images, despite the datasets from which they were selected clearly demonstrating a drastic change in FID. 
    The FID, WaM, and $R$ values were calculated using \inception{}.
    }
    \label{fig:perturbed noise}
\end{figure*}

\subsection{Targeted perturbations --- WaM captures more information than FID}
\label{sec:targeted perturbations}
The purpose of this experiment is to show that WaM can capture more information than FID by implicitly capturing higher order moments.
Although features extracted from classifiers are not Gaussian, we do not have a perfect model for them. In fact, it is difficult to come up with distributions of features without images to start with. 
Thus, we start with a set of images, perturb them in order to change their first and second moments, then calculate WaM and FID on the perturbed images. Since WaM is a generalization of FID, the perturbed images will likely affect both WaM and FID. However, since WaM can capture more information than FID on the feature distributions, we hypothesize that WaM will not be as affected as FID.

We construct these perturbed sets of images by trying to \textit{maximize} the following losses
\begin{align}
    \mcL(\vmu) &= \|\vmu - \vmu_0\|_2^2
    \label{eq:meanPert}
    \\
    \mcL(\mSig) &= \|\mSig - \mSig_0\|_F
    \label{eq:covPert}
    \\
    \mcL(\vmu, \mSig) &= \frac{1}{2}\Big(\|\vmu - \vmu_0\|_2^2 + \|\mSig - \mSig_0\|_F \Big)
    \label{eq:meancovPert}
    \\
    \mcL(\mSig) &= \trace(\mSig) + \trace(\mSig_0) - 2 \trace\Big(  \big(\mSig^\frac{1}{2} \mSig_0 \mSig^\frac{1}{2}\big)^\frac{1}{2} \Big)
    \label{eq:W2CovPert}
\end{align}
using the Fast Gradient Sign Method (FGSM) \cite{goodfellow2014explaining}, where $\vmu_0$ and $\mSig_0$ are the fixed first and second moment of the \imagenet{} training data. In addition we adversarially perturb FID and report our findings in more detail in \cref{app:targeted}. In \cref{fig:perturbed noise} we show how the mean perturbation using \cref{eq:meanPert} affects FID significantly more than WaM even though there are no visual differences.

\begin{figure*}
    \centering
    \def\figWidth{0.16}

    \setlength\tabcolsep{1pt}
    \begin{tabular}{lccccccc}%
        \multicolumn{1}{c}{original} & $\sigma = 0.01$ & $\sigma = 0.05$ & $\sigma = 0.1$ & $\sigma = 0.2$ & $\sigma = 0.5$ \\
        \includegraphics[width=\figWidth\textwidth]{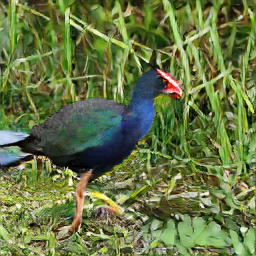}
        &
        \includegraphics[width=\figWidth\textwidth]{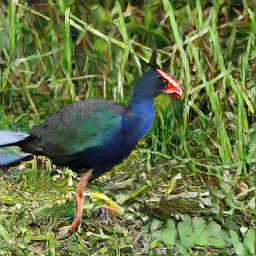}
        &
        \includegraphics[width=\figWidth\textwidth]{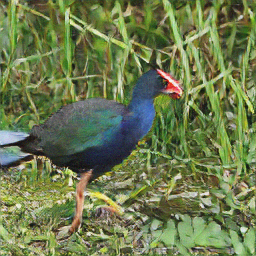}
        &
        \includegraphics[width=\figWidth\textwidth]{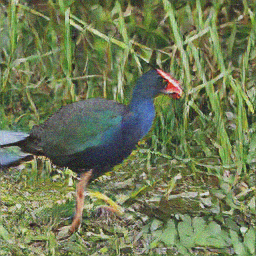}
        &
        \includegraphics[width=\figWidth\textwidth]{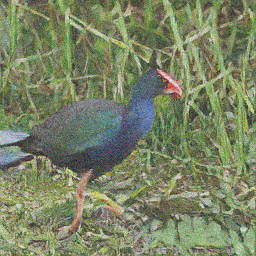}
        &
        \includegraphics[width=\figWidth\textwidth]{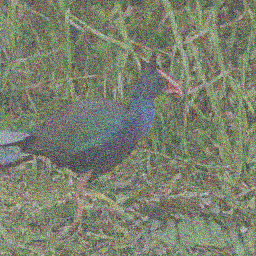}
        
        \ \\
        \\
        \toprule 
        & $\sigma = 0.01$ & $\sigma = 0.05$ & $\sigma = 0.1$ & $\sigma = 0.2$ & $\sigma = 0.5$\\ \toprule
        \FID(orig) & 24.14 & 24.14 & 24.14 & 24.14 & 24.14\\
        \FID(pert) & 24.37 & 27.10 & 33.55 & 51.10 & 114.94\\
        \WaM$^2$(orig) & 504.30 & 504.30 & 504.30 & 504.30 & 504.30\\
        \WaM$^2$(pert) & 539.54 &  516.75 & 628.68 & 748.65 & \num{1328.01} \\ \midrule 
        $R_\FID$ & 1.01 & 1.12 & 1.39 & 2.12 & 4.76 \\
        $R_\WaM$ & 1.07 & 1.02 & 1.25 & 1.48 & 2.63 \\
        $R$      & \textbf{0.94} & \textbf{1.10} & \textbf{1.11} & \textbf{1.43} & \textbf{1.81} \\
        \bottomrule
    \end{tabular}
    
    \caption{
    $R$ values for BigGAN-generated images using additive isotropic Gaussian noise showing that FID is slightly more sensitive than WaM to noise perturbations of generated images. 
    The noise perturbations in this experiment are all greater in magnitude than the targeted perturbations in \cref{sec:targeted perturbations}. 
    The original image above was randomly selected from a set of \num{50000} images generated by BigGAN.
    The FID, WaM, and $R$ values were calculated using \resnetE{}.}
    \label{fig:BigGAN random noise}
\end{figure*}

\begin{figure*}
    \centering
    \def\figWidth{0.16}
    \setlength\tabcolsep{1pt}
    \begin{tabular}{lccccc}
        \multicolumn{1}{c}{original} & $\sigma = 0.01$ & $\sigma = 0.05$ & $\sigma = 0.1$ & $\sigma = 0.2$ & $\sigma = 0.5$ \\
        \includegraphics[width=\figWidth\textwidth]{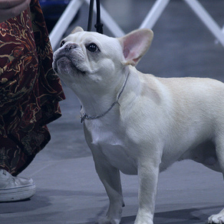}
        &
        \includegraphics[width=\figWidth\textwidth]{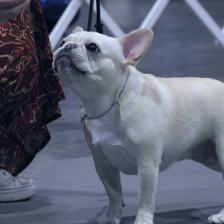}
        &
        \includegraphics[width=\figWidth\textwidth]{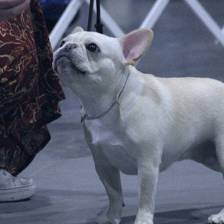}
        &
        \includegraphics[width=\figWidth\textwidth]{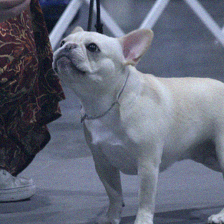}
        &
        \includegraphics[width=\figWidth\textwidth]{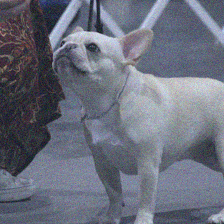}
        &
        \includegraphics[width=\figWidth\textwidth]{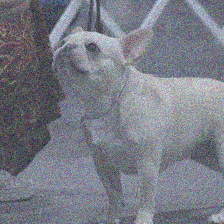}
        \\ \ \\
        \toprule
        & $\sigma = 0.01$ & $\sigma = 0.05$ & $\sigma = 0.1$ & $\sigma = 0.2$ & $\sigma = 0.5$\\ \toprule  
        \FID(orig) & 3.61 & 3.61 & 3.61  & 3.61 & 3.61 \\
        \FID(pert) & 5.07 & 21.79 & 52.30 & 120.05 & 322.84 \\
        \WaM$^2$(orig) & 208.45  & 208.45  & 208.45  & 208.45  & 208.45\\
        \WaM$^2$(pert) & 219.49 & 316.06 & 549.03 & 1081.28 & 4007.29\\ \midrule  
        $R_\FID$ & 1.41 & 6.04 & 14.49 & 33.26 & 89.45 \\
        $R_\WaM$ & 1.05 & 1.52 & 2.63 & 5.19 & 19.22 \\
        $R$      & \textbf{1.34} &
        \textbf{3.98} & \textbf{5.50} & \textbf{6.41} & \textbf{4.65} 
        \\
        \bottomrule
    \end{tabular}

    \caption{$R$ values for real images (\imagenet{} validation data) using additive isotropic Gaussian noise showing that FID is significantly more sensitive than WaM to noise perturbations of real images. 
    The noise perturbations in this experiment are all greater in magnitude than the targeted perturbations in \cref{sec:targeted perturbations}. 
    The original image above was randomly selected from a set of \num{50000} images of the \imagenet{} validation dataset.
    In contrast to \cref{fig:BigGAN random noise}, we see that FID is more sensitive to these perturbations when the images look more realistic.
    The FID and WaM values were calculated using \resnetE{}.
    }
    \label{fig:ImagetNet random noise}
\end{figure*}

To calculate FID or WaM, we must compare two sets of images; thus, we always compare to the \imagenet{} training set~\cite{biggan}. This allows us to calculate the FID and WaM of the \imagenet{} training set against real images from the \imagenet{} validation set, generated images from BigGAN~\cite{biggan}, and perturbed images from each. 
We compare to real images because we want our metrics to work well with the most realistic images possible,  given the continuously improving nature of GANs.
We used \num{50000} images for  doing all the comparisons and the whole training set for the reference. To produce the adversarial images, we extracted the features from all the \num{50000} \imagenet{} validation images, then ran FGSM with an $\epsilon=0.01$ and batch size of $64$ until we perturbed all \num{50000} of our target images (e.g., \imagenet{} validation set). This means that the maximum difference per pixel is $0.25\%$. During training we calculated the gradients that maximize the losses above between the features of a batch of $64$ images and the features of the \imagenet{} training set.

Comparing FID and WaM is difficult because they are different metrics with different scales. For this reason, we must normalize them when comparing. Thus, we define $R_\FID$ to be the ratio of the FID of the perturbed images over the FID of the original images. Hence, $R_\FID$ shows how much FID has increased due to the perturbation. Similarly, we define $R_\WaM$ to be the ratio of WaM squared of the perturbed images over WaM squared of the original images. FID is typically reported as the $2$-Wasserstein squared distance, so we square WaM so that it is also a squared distance. Then we define $R = \frac{R_\FID}{R_\WaM}$ to be the ratio for these two increases. Thus, for $R > 1$ we have that FID increased faster than WaM due to perturbation.

When we perturb images generated from BigGAN~\cite{biggan} or the \imagenet{} training data we cannot visually perceive a difference, as shown in \cref{fig:perturbed noise}. However, for the BigGAN images, FID increases by a factor of $R_\FID = 2.77$ while WaM only increases by a factor of $R_\WaM = 1.12$.
This difference is even significantly more evident with real images drawn from the \imagenet{} training data set. We see that the FID score after perturbation increases by $R_\FID = 12.74$ times. Since WaM only increases by $R_\WaM = 1.18$ times, we see that FID increased $R = 10.78$ times more than WaM for an imperceptible, but targeted, perturbation. That is an extremely large sensitivity to noise that human eyes cannot even see.
A metric which reflects perceptual quality perfectly would not be affected whatsoever by these perturbations. Neither FID nor WaM are perfect, but WaM's lower sensitivity to visually imperceptible perturbation is better aligned with the objective of assessing perceptual quality in images.

Even though these perturbations are targeted to specifically change the first two moments of the data, we note that WaM is still affected by these perturbations. This is because WaM can capture more moments of the data than FID. More specifically, WaM can learn a Gaussian distribution (e.g., if all the components are the same), yet FID and WaM yield different results in this experiment, implying that the features are not modeled well by FID and benefit from the additional information captured by WaM.

\subsection{Random perturbations}
\label{sec:random pert}

In this section we show that WaM is also less sensitive than FID to additive isotropic Gaussian noise. We do this by corrupting images generated from BigGAN and the \imagenet{} training dataset by adding isotropic Gaussian noise with standard deviation $\sigma \in \{0.01, 0.05, 0.1, 0.2, 0.5\}$ and then calculating their features. Samples of how these noisy images compare to the original are shown in \cref{fig:BigGAN random noise,fig:ImagetNet random noise}. In these experiments, we use \resnetE{} to extract the features. The $\epsilon = 0.01$ used in \cref{sec:targeted perturbations} corresponds to approximately $\sigma = 0.0014$, meaning that the additive random noise in \cref{fig:BigGAN random noise,fig:ImagetNet random noise} perturbs the images much more than the targeted noise in \cref{fig:perturbed noise}.

We see that FID and WaM perform similarly when calculated using noisy BigGAN generated images, but WaM is still significantly more robust than FID (see \cref{fig:BigGAN random noise}). Moreover, FID skyrockets when calculated using \imagenet{} training data. This is likely due to FID not being able to capture the differences between the \imagenet{} training and validation set. One can justly assume that both data sets are sampled from the same distribution; however, we are not comparing the distributions from which they are sampled. We are comparing the two sets of images from the training and validation set, which are not the same. Therefore, FID's inability to model the correct distribution of features causes it to become extremely sensitive to this noise, even when it is barely visually perceptible. 
This sensitivity of FID to noise has been noted before~\cite{FID,borji2019pros}.
FID is affected $R=5.50$ times as much as WaM when the noise is barely visible ($\sigma = 0.1$), making WaM much more desirable to use in noisy contexts (see \cref{fig:ImagetNet random noise}).

A good metric for evaluating generative network performance should be able to capture the quality of generated images at all stages. FID does not do this well. FID is sensitive to noise perturbations, especially when the images look realistic; hence, $R$ is much larger for the \imagenet{} training data than it is for the BigGAN generated images. As generative networks get better and better, we must use more information (not just the first and second moment) from the feature distribution in order to accurately evaluate generated samples.

\section{Conclusions }
\label{sec:Conclusion}

We generalize the notion of FID by modeling image features with GMMs and computing a relaxed $2$-Wasserstein distance on the distributions.
Our proposed metric, WaM, allows us to accurately model more complex distributions than FID, which relies on the invalid assumption that image features follow a Gaussian distribution. 
Moreover, we show that WaM is less sensitive to both imperceptible targeted perturbations that modify the first two moments of the feature distribution and imperceptible additive Gaussian noise. This is important because we want a performance metric which
is truly reflective of the perceptual quality of images and 
will not vary much when visually imperceptible noise is added. 
We can use WaM to evaluate networks which generate new images, superresolve images, solve inverse problems, perform image-to-image translation tasks, and more. As networks continue to evolve and generate more realistic images, WaM can provide a superior model of the feature distributions, thus enabling more accurate evaluation of extremely-realistic generated images.

\newpage
\section*{Acknowledgements}
Rice University affiliates were partially supported by NSF grants CCF-1911094, IIS-1838177, and IIS-1730574; ONR grants N00014-18-12571, N00014-20-1-2534, and MURI N00014-20-1-2787; AFOSR grant FA9550-18-1-0478; and a Vannevar Bush Faculty Fellowship, ONR grant N00014-18-1-2047.

{\small
\bibliographystyle{ieee_fullname}
\bibliography{egbib}
}

\clearpage
\clearpage
\clearpage
\appendix
\section{Targeted perturbations (extended)}
\label{app:targeted}

Here we show the targeted perturbation results in more detail than in \cref{sec:targeted perturbations}. We don't show figures of the images before and after perturbation besides \cref{fig:perturbed noise} because they are all imperceptible, with maximum pixel differences of $0.25\%$. All the FID, WaM, and $R$ values were calculated using \inception{}. We use \cref{eq:meanPert} for \cref{fig:meanPertFID}, \cref{eq:covPert} for \cref{fig:covPertFID}, \cref{eq:meancovPert} for \cref{fig:meancovPertFID}, \cref{eq:W2CovPert} for \cref{fig:W2covPertFID}, and FID for \cref{fig:FIDPertFID}. We follow recent work~\cite{mathiasen2020fast,mathiasen2020backpropagating}\footnote{Although the authors of the paper introduced a Fast FID, we backpropagate through FID in our work.} in order to backpropagate through FID. Our results show that in every case, FID is significantly more sensitive to imperceptible perturbations of the first two moments when compared to WaM.

\begin{table*}[!hb]
    \centering
    \def\figWidth{0.24}
    \bgroup
    \begin{tabular}{cc|cc}
        \toprule
        \textbf{Original (BigGAN)}
        &
        \textbf{Perturbed (BigGAN)}
        &
        \textbf{Original (ImageNet)}
        &
        \textbf{Perturbed (ImageNet)}
        \\ \midrule
        $\!\begin{aligned}
        \FID &= 55.7\\
        \WaM^2 &= 378.4
        \end{aligned}$
        &
        $\!\begin{aligned}
        \FID &= 154.2\\
        \WaM^2 &= 424.3
        \end{aligned}$
        &
        $\!\begin{aligned}
        \FID &= 3.7\\
        \WaM^2 &= 237.0
        \end{aligned}$
        &
        $\!\begin{aligned}
        \FID &= 46.6 \\
        \WaM^2 &= 280.0 
        \end{aligned}$
        \\ \midrule
        \multicolumn{2}{c|}
        {
        $\!\begin{aligned}
        R_\FID &= 2.8 \\
        R_\WaM &= 1.1 \\
        R &= \mathbf{ 2.5 }\end{aligned}$}
        &
        \multicolumn{2}{c}
        {$\!\begin{aligned}
        R_\FID &= 12.7 \\
        R_\WaM &= 1.2 \\
        R &= \mathbf{ 10.8 }\end{aligned}$}
        \\ \bottomrule
    \end{tabular}
    \egroup
    \caption{ \textbf{Mean perturbations}: We show that FID values are significantly more sensitive to imperceptible perturbations to the feature means (\cref{eq:meanPert}).
    }
    \label{fig:meanPertFID}
\end{table*}

\begin{table*}[!hb]
    \centering
    \def\figWidth{0.24}
    \bgroup
    \begin{tabular}{cc|cc}
        \toprule
        \textbf{Original (BigGAN)}
        &
        \textbf{Perturbed (BigGAN)}
        &
        \textbf{Original (ImageNet)}
        &
        \textbf{Perturbed (ImageNet)}
        \\ \midrule
        $\!\begin{aligned}
        \FID &= 55.7\\
        \WaM^2 &= 378.4
        \end{aligned}$
        &
        $\!\begin{aligned}
        \FID &= 125.7\\
        \WaM^2 &= 540.8
        \end{aligned}$
        &
        $\!\begin{aligned}
        \FID &= 3.7\\
        \WaM^2 &= 237.0
        \end{aligned}$
        &
        $\!\begin{aligned}
        \FID &= 113.0 \\
        \WaM^2 &= 422.5 
        \end{aligned}$
        \\ \midrule
        \multicolumn{2}{c|}
        {
        $\!\begin{aligned}
        R_\FID &= 2.3 \\
        R_\WaM &= 1.4 \\
        R &= \mathbf{ 1.6 }\end{aligned}$}
        &
        \multicolumn{2}{c}
        {$\!\begin{aligned}
        R_\FID &= 30.9 \\
        R_\WaM &= 1.8 \\
        R &= \mathbf{ 17.3 }\end{aligned}$}
        \\ \bottomrule
    \end{tabular}
    \egroup
    \caption{ \textbf{Covariance perturbations}: We show that FID values are significantly more sensitive to imperceptible perturbations to the feature covariances (\cref{eq:covPert}).
    }
    \label{fig:covPertFID}
\end{table*}

\begin{table*}[!hb]
    \centering
    \def\figWidth{0.24}
    \bgroup
    \begin{tabular}{cc|cc}
        \toprule
        \textbf{Original (BigGAN)}
        &
        \textbf{Perturbed (BigGAN)}
        &
        \textbf{Original (ImageNet)}
        &
        \textbf{Perturbed (ImageNet)}
        \\ \midrule
        $\!\begin{aligned}
        \FID &= 55.7\\
        \WaM^2 &= 378.4
        \end{aligned}$
        &
        $\!\begin{aligned}
        \FID &= 177.6 \\
        \WaM^2 &= 521.3
        \end{aligned}$
        &
        $\!\begin{aligned}
        \FID &= 3.7\\
        \WaM^2 &= 237.0
        \end{aligned}$
        &
        $\!\begin{aligned}
        \FID &= 106.9 \\
        \WaM^2 &= 412.2 
        \end{aligned}$
        \\ \midrule
        \multicolumn{2}{c|}
        {
        $\!\begin{aligned}
        R_\FID &= 3.2 \\
        R_\WaM &= 1.4 \\
        R &= \mathbf{ 2.3 }\end{aligned}$}
        &
        \multicolumn{2}{c}
        {$\!\begin{aligned}
        R_\FID &= 29.2 \\
        R_\WaM &= 1.7 \\
        R &= \mathbf{ 16.8 }\end{aligned}$}
        \\ \bottomrule
    \end{tabular}
    \egroup
    \caption{ \textbf{Mean-covariance perturbations}: We show that FID values are significantly more sensitive to imperceptible perturbations to the feature means and covariances together (\cref{eq:meancovPert}).
    }
    \label{fig:meancovPertFID}
\end{table*}

\begin{table*}[!hb]
    \centering
    \def\figWidth{0.24}
    \bgroup
    \begin{tabular}{cc|cc}
        \toprule
        \textbf{Original (BigGAN)}
        &
        \textbf{Perturbed (BigGAN)}
        &
        \textbf{Original (ImageNet)}
        &
        \textbf{Perturbed (ImageNet)}
        \\ \midrule
        $\!\begin{aligned}
        \FID &= 55.7\\
        \WaM^2 &= 378.4
        \end{aligned}$
        &
        $\!\begin{aligned}
        \FID &= 145.9 \\
        \WaM^2 &= 578.3
        \end{aligned}$
        &
        $\!\begin{aligned}
        \FID &= 3.7\\
        \WaM^2 &= 237.0
        \end{aligned}$
        &
        $\!\begin{aligned}
        \FID &= 112.2 \\
        \WaM^2 &= 444.0 
        \end{aligned}$
        \\ \midrule
        \multicolumn{2}{c|}
        {
        $\!\begin{aligned}
        R_\FID &= 2.6 \\
        R_\WaM &= 1.5 \\
        R &= \mathbf{ 1.7 }\end{aligned}$}
        &
        \multicolumn{2}{c}
        {$\!\begin{aligned}
        R_\FID &= 30.7 \\
        R_\WaM &= 1.9 \\
        R &= \mathbf{ 16.4 }\end{aligned}$}
        \\ \bottomrule
    \end{tabular}
    \egroup
    \caption{ \textbf{Alternative covariance perturbations}: We show that FID values are significantly more sensitive to imperceptible perturbations to the feature covariances, using a different metric on the covariances than the Frobenius norm (\cref{eq:W2CovPert}).
    }
    \label{fig:W2covPertFID}
\end{table*}

\begin{table*}[!hb]
    \centering
    \def\figWidth{0.24}
    \bgroup
    \begin{tabular}{cc|cc}
        \toprule
        \textbf{Original (BigGAN)}
        &
        \textbf{Perturbed (BigGAN)}
        &
        \textbf{Original (ImageNet)}
        &
        \textbf{Perturbed (ImageNet)}
        \\ \midrule
        $\!\begin{aligned}
        \FID &= 55.7\\
        \WaM^2 &= 378.4
        \end{aligned}$
        &
        $\!\begin{aligned}
        \FID &= 166.5 \\
        \WaM^2 &= 548.5
        \end{aligned}$
        &
        $\!\begin{aligned}
        \FID &= 3.7\\
        \WaM^2 &= 237.0
        \end{aligned}$
        &
        $\!\begin{aligned}
        \FID &= 112.0 \\
        \WaM^2 &= 377.0 
        \end{aligned}$
        \\ \midrule
        \multicolumn{2}{c|}
        {
        $\!\begin{aligned}
        R_\FID &= 3.0 \\
        R_\WaM &= 1.4 \\
        R &= \mathbf{ 2.1 }\end{aligned}$}
        &
        \multicolumn{2}{c}
        {$\!\begin{aligned}
        R_\FID &= 30.6 \\
        R_\WaM &= 1.6 \\
        R &= \mathbf{ 19.2 }\end{aligned}$}
        \\ \bottomrule
    \end{tabular}
    \egroup
    \caption{ \textbf{FID perturbations}: We show that FID values are significantly more sensitive to imperceptible perturbations when we adversarially attempt to inflate FID.
    }
    \label{fig:FIDPertFID}
\end{table*}

\section{Kernel Inception distance experiments}
\label{app:KID}
Kernel Inception distance (KID)~\cite{binkowski2018demystifying} is a popular method to evaluate the performance of a GAN which uses embeddings from powerful classifiers, such as \inception{}~\cite{szegedy2016rethinking}. We use the cubic polynomial kernel, i.e., $k(\vx, \vy) = (\frac{1}{d}\vx^\transp \vy + 1)^3$ for $\vx, \vy \in \bbR^d$, to compute similarities between featurized samples, as is typically done. We use this method to evaluate WaM's sensitivity to imperceptible noise perturbations. To do this, we define $R_{KID}$ to be the ratio of the
KID of the perturbed images over the KID of the original images. We further define 
\[
    R' = \frac{R_\text{KID}}{R_\text{WaM}}.
\]
All the KID, WaM, and $R$ values were calculated using \inception{}. We use \cref{eq:meanPert} for \cref{fig:meanPertKID}, \cref{eq:covPert} for \cref{fig:covPertKID}, \cref{eq:meancovPert} for \cref{fig:meancovPertKID}, \cref{eq:W2CovPert} for \cref{fig:W2covPertKID}, and FID for \cref{fig:FIDPertKID}.
These results show that KID is is still significantly affected by these perturbations, even though some values of $R_\KID$ are smaller than $R_\WaM$. WaM is less sensitive than both FID and KID in the majority of these experiments, implying that it does not depend as heavily on the first two moments and can capture more higher order information than both metrics.

\begin{table*}[!hb]
    \centering
    \def\figWidth{0.24}
    \bgroup
    \begin{tabular}{cc|cc}
        \toprule
        \textbf{Original (BigGAN)}
        &
        \textbf{Perturbed (BigGAN)}
        &
        \textbf{Original (ImageNet)}
        &
        \textbf{Perturbed (ImageNet)}
        \\ \midrule
        $\!\begin{aligned}
        \KID &= 0.029\\
        \WaM^2 &= 378.4
        \end{aligned}$
        &
        $\!\begin{aligned}
        \KID &= 0.139 \\
        \WaM^2 &= 424.3
        \end{aligned}$
        &
        $\!\begin{aligned}
        \KID &= 0.0007\\
        \WaM^2 &= 237.0
        \end{aligned}$
        &
        $\!\begin{aligned}
        \KID &= 0.066 \\
        \WaM^2 &= 280.0 
        \end{aligned}$
        \\ \midrule
        \multicolumn{2}{c|}
        {
        $\!\begin{aligned}
        R_\KID &= 4.7 \\
        R_\WaM &= 1.1 \\
        R' &= \mathbf{ 4.2 }\end{aligned}$}
        &
        \multicolumn{2}{c}
        {$\!\begin{aligned}
        R_\KID &= 94.6 \\
        R_\WaM &= 1.2 \\
        R' &= \mathbf{ 80.1 }\end{aligned}$}
        \\ \bottomrule
    \end{tabular}
    \egroup
    \caption{ \textbf{Mean perturbations}: We show that KID values are significantly more sensitive to imperceptible perturbations to the feature means (\cref{eq:meanPert}).
    }
    \label{fig:meanPertKID}
\end{table*}

\begin{table*}[!hb]
    \centering
    \def\figWidth{0.24}
    \bgroup
    \begin{tabular}{cc|cc}
        \toprule
        \textbf{Original (BigGAN)}
        &
        \textbf{Perturbed (BigGAN)}
        &
        \textbf{Original (ImageNet)}
        &
        \textbf{Perturbed (ImageNet)}
        \\ \midrule
        $\!\begin{aligned}
        \KID &= 0.029\\
        \WaM^2 &= 378.4
        \end{aligned}$
        &
        $\!\begin{aligned}
        \KID &= 0.014 \\
        \WaM^2 &= 540.8
        \end{aligned}$
        &
        $\!\begin{aligned}
        \KID &= 0.0007 \\
        \WaM^2 &= 237.0
        \end{aligned}$
        &
        $\!\begin{aligned}
        \KID &= 0.087 \\
        \WaM^2 &= 422.5 
        \end{aligned}$
        \\ \midrule
        \multicolumn{2}{c|}
        {
        $\!\begin{aligned}
        R_\KID &= 0.5 \\
        R_\WaM &= 1.4 \\
        R' &= \mathbf{ 0.3 }\end{aligned}$}
        &
        \multicolumn{2}{c}
        {$\!\begin{aligned}
        R_\KID &= 125.6 \\
        R_\WaM &= 1.8 \\
        R' &= \mathbf{ 70.5 }\end{aligned}$}
        \\ \bottomrule
    \end{tabular}
    \egroup
    \caption{  \textbf{Covariance perturbations}: We show that KID values are significantly more sensitive to imperceptible perturbations to the feature covariances (\cref{eq:covPert}).
    }
    \label{fig:covPertKID}
\end{table*}

\begin{table*}[!hb]
    \centering
    \def\figWidth{0.24}
    \bgroup
    \begin{tabular}{cc|cc}
        \toprule
        \textbf{Original (BigGAN)}
        &
        \textbf{Perturbed (BigGAN)}
        &
        \textbf{Original (ImageNet)}
        &
        \textbf{Perturbed (ImageNet)}
        \\ \midrule
        $\!\begin{aligned}
        \KID &= 0.029 \\
        \WaM^2 &= 378.4
        \end{aligned}$
        &
        $\!\begin{aligned}
        \KID &= 0.097 \\
        \WaM^2 &= 521.3
        \end{aligned}$
        &
        $\!\begin{aligned}
        \KID &= 0.0007 \\
        \WaM^2 &= 237.0
        \end{aligned}$
        &
        $\!\begin{aligned}
        \KID &= 0.100 \\
        \WaM^2 &= 412.2 
        \end{aligned}$
        \\ \midrule
        \multicolumn{2}{c|}
        {
        $\!\begin{aligned}
        R_\KID &= 3.3 \\
        R_\WaM &= 1.4 \\
        R' &= \mathbf{ 2.4 }\end{aligned}$}
        &
        \multicolumn{2}{c}
        {$\!\begin{aligned}
        R_\KID &= 143.9 \\
        R_\WaM &= 1.7 \\
        R' &= \mathbf{ 80.8 }\end{aligned}$}
        \\ \bottomrule
    \end{tabular}
    \egroup
    \caption{ \textbf{Mean-covariance perturbations}: We show that KID values are significantly more sensitive to imperceptible perturbations to the feature means and covariances together (\cref{eq:meancovPert}).
    }
    \label{fig:meancovPertKID}
\end{table*}

\begin{table*}[!hb]
    \centering
    \def\figWidth{0.24}
    \bgroup
    \begin{tabular}{cc|cc}
        \toprule
        \textbf{Original (BigGAN)}
        &
        \textbf{Perturbed (BigGAN)}
        &
        \textbf{Original (ImageNet)}
        &
        \textbf{Perturbed (ImageNet)}
        \\ \midrule
        $\!\begin{aligned}
        \KID &= 0.029 \\
        \WaM^2 &= 378.4
        \end{aligned}$
        &
        $\!\begin{aligned}
        \KID &= 0.034 \\
        \WaM^2 &= 578.3
        \end{aligned}$
        &
        $\!\begin{aligned}
        \KID &= 0.0007 \\
        \WaM^2 &= 237.0
        \end{aligned}$
        &
        $\!\begin{aligned}
        \KID &= 0.074 \\
        \WaM^2 &= 444.0 
        \end{aligned}$
        \\ \midrule
        \multicolumn{2}{c|}
        {
        $\!\begin{aligned}
        R_\KID &= 1.2 \\
        R_\WaM &= 1.5 \\
        R' &= \mathbf{ 0.8 }\end{aligned}$}
        &
        \multicolumn{2}{c}
        {$\!\begin{aligned}
        R_\KID &= 106.1 \\
        R_\WaM &= 1.9 \\
        R' &= \mathbf{ 56.6 }\end{aligned}$}
        \\ \bottomrule
    \end{tabular}
    \egroup
    \caption{\textbf{Alternative covariance perturbations}: We show that KID values are significantly more sensitive to imperceptible perturbations to the feature covariances, using a different metric on the covariances than the Frobenius norm (\cref{eq:W2CovPert}).
    }
    \label{fig:W2covPertKID}
\end{table*}

\begin{table*}[!hb]
    \centering
    \def\figWidth{0.24}
    \bgroup
    \begin{tabular}{cc|cc}
        \toprule
        \textbf{Original (BigGAN)}
        &
        \textbf{Perturbed (BigGAN)}
        &
        \textbf{Original (ImageNet)}
        &
        \textbf{Perturbed (ImageNet)}
        \\ \midrule
        $\!\begin{aligned}
        \KID &= 0.029 \\
        \WaM^2 &= 378.4
        \end{aligned}$
        &
        $\!\begin{aligned}
        \KID &= 0.057 \\
        \WaM^2 &= 548.5
        \end{aligned}$
        &
        $\!\begin{aligned}
        \KID &= 0.0007 \\
        \WaM^2 &= 237.0
        \end{aligned}$
        &
        $\!\begin{aligned}
        \KID &= 0.077 \\
        \WaM^2 &= 377.0 
        \end{aligned}$
        \\ \midrule
        \multicolumn{2}{c|}
        {
        $\!\begin{aligned}
        R_\KID &= 2.0 \\
        R_\WaM &= 1.4 \\
        R' &= \mathbf{ 1.3 }\end{aligned}$}
        &
        \multicolumn{2}{c}
        {$\!\begin{aligned}
        R_\KID &= 111.5 \\
        R_\WaM &= 1.6 \\
        R' &= \mathbf{ 70.1 }\end{aligned}$}
        \\ \bottomrule
    \end{tabular}
    \egroup
    \caption{ \textbf{FID perturbations}: We show KID values are significantly more sensitive to imperceptible perturbations when we adversarially attempt to inflate FID.
    }
    \label{fig:FIDPertKID}
\end{table*}

We now consider the random perturbations in \cref{sec:random pert} of the original paper and evaluate $R'$ on them, as shown in \cref{fig:BigGAN random noiseKID,fig:ImagetNet random noiseKID}. We see that KID has similar sensitivity to WaM on BigGAN generated images but much higher sensitivity on real images. In fact, KID has higher sensitivity on real images than FID. We stress that the ability to evaluate realistic images is important because that is what we \textbf{want} to generate. Therefore, WaM provides a means to evaluate realistic images better than FID and KID under imperceptible noise perturbations.

\begin{table*}[!hb]
    \centering
    \bgroup
    \setlength\tabcolsep{12pt}
    \begin{tabular}{lccccccc}%
    \toprule
        & $\sigma = 0.01$ & $\sigma = 0.05$ & $\sigma = 0.1$ & $\sigma = 0.2$ & $\sigma = 0.5$\\ \toprule
        \KID(orig) & 2.11 &    2.11 &   2.11    & 2.11 &   2.11 \\
        \KID(pert) & 2.22 &    2.74  &  3.47    &4.65 &   5.23 \\
        \WaM$^2$(orig) & 504.30 & 504.30 & 504.30 & 504.30 & 504.30\\
        \WaM$^2$(pert) & 539.54 &  516.75 & 628.68 & 748.65 & \num{1328.01}\\ \midrule 
        $R_\KID$ & 1.05 &   1.29&    1.64&    2.2&     2.47   \\
        $R_\WaM$ & 1.07 & 1.02 & 1.25 & 1.48 & 2.63 \\
        $R'$      & \textbf{0.98}  &  \textbf{1.26}  &  \textbf{1.31} &   \textbf{1.49}  &  \textbf{0.94}
        \\ \bottomrule
        
    \end{tabular}
    \egroup
    
    \caption{
    $R'$ values for BigGAN-generated images using additive isotropic Gaussian noise (as explained in \cref{sec:random pert}) showing that KID has similar sensitivity as WaM to noise perturbations of generated images. 
    The original image above was randomly selected from a set of \num{50000} images generated by BigGAN.
    The KID, WaM, and $R'$ values were calculated using \resnetE{}.}
    \label{fig:BigGAN random noiseKID}
\end{table*}

\begin{table*}[t]
    \centering
    \setlength\tabcolsep{12pt}
    \begin{tabular}{lccccccc}%
        \toprule 
        & $\sigma = 0.01$ & $\sigma = 0.05$ & $\sigma = 0.1$ & $\sigma = 0.2$ & $\sigma = 0.5$\\ \toprule 
        \KID(orig) & 0.025 &  0.025 &  0.025 &  0.025 &  0.025 \\
        \KID(pert) & 0.033   & 0.187 &  0.496 &  1.146 &  2.745  \\
        \WaM$^2$(orig) & 208.45  & 208.45  & 208.45  & 208.45  & 208.45\\
        \WaM$^2$(pert) & 219.49 & 316.06 & 549.03 & 1081.28 & 4007.29\\ \midrule  
        $R_\KID$ & 1.283 &   7.363  & 19.528 & 45.143 & 108.118\\
        $R_\WaM$ & 1.05 & 1.52 & 2.63 & 5.19 & 19.22 \\
        $R'$      & \textbf{1.22} &  \textbf{4.84} &  \textbf{7.43} &  \textbf{8.70} & \textbf{5.63}  
        \\ \bottomrule
    \end{tabular}

    \caption{$R'$ values for real images (\imagenet{} validation data) using additive isotropic Gaussian noise (as explained in \cref{sec:random pert}) showing that KID is more sensitive than WaM to noise perturbations of real images. 
    The original image above was randomly selected from a set of \num{50000} images of the \imagenet{} validation dataset.
    In contrast to \cref{fig:BigGAN random noise}, we see that KID is more sensitive to these perturbations when the images look more realistic.
    The FID and WaM values were calculated using \resnetE{}.
    }
    \label{fig:ImagetNet random noiseKID}
\end{table*}

\end{document}